%% file: main.tex
\documentclass[letterpaper, 10 pt, conference]{ieeeconf}  
\usepackage{floatrow}
\newfloatcommand{capbtabbox}{table}[][\FBwidth]

\usepackage{wrapfig}
\usepackage{amssymb}
\usepackage{times}
\usepackage{caption}
\usepackage[labelformat=simple]{subcaption}
\usepackage{graphicx}
\usepackage{graphics}
\usepackage{blkarray}
\usepackage{booktabs}
\usepackage{xcolor} 
\usepackage{bbold}
\usepackage{caption}
\usepackage{subcaption}
\usepackage{multirow}
\usepackage{comment}
\usepackage{epstopdf}
\usepackage{tikz}
\usepackage{todonotes}
\graphicspath{{./figures/}}
\usepackage{csquotes}
\usepackage{siunitx}
\usepackage{cite}
\usepackage{epigraph}
\usepackage{empheq}
\usepackage{comment}
\usepackage{cancel}
\usepackage{algorithm} 
\usepackage{algorithmic}  
\usepackage{mathtools}
\usepackage{dsfont}

\usepackage{balance}
\usepackage{dirtytalk}

\usepackage[linesnumbered,ruled,vlined,algo2e,noend]{algorithm2e}

\SetInd{0.35em}{0.5em}

\usepackage{hyperref}

\newcommand{\onte}[1]{\textit{#1}}
\newcommand{\ote}[1]{\textbf{#1}}

\newcommand{\btrun}{\textsf{running}}

\SetKwRepeat{Do}{do}{while}

\makeatother

\DeclareCaptionLabelSeparator{periodspace}{.\quad}
\IEEEoverridecommandlockouts                              
\newtheorem{experiment}{Experiment}

\makeatletter
\newcommand*{\rom}[1]{\expandafter\@slowromancap\romannumeral #1@}
\makeatother

\def\eqalignno#1{\let\\=\cr\displ@y \tabskip\@centering
  \halign to\displaywidth{\hfil$\@lign\displaystyle{##}$\tabskip\z@skip
    &$\@lign\displaystyle{{}##}$\hfil\tabskip\@centering
    &\llap{$\@lign##$}\tabskip\z@skip\crcr
    #1\crcr}}
\def\leqalignno#1{\let\\=\cr\displ@y \tabskip\@centering
  \halign to\displaywidth{\hfil$\@lign\displaystyle{##}$\tabskip\z@skip
    &$\@lign\displaystyle{{}##}$\hfil\tabskip\@centering
    &\kern-\displaywidth\rlap{$\@lign##$}\tabskip\displaywidth\crcr
    #1\crcr}}

\begin{document}

\thispagestyle{empty}
\twocolumn
\title{\LARGE \bf
              Formalizing the Execution Context of Behavior Trees\\
              for Runtime Verification of Deliberative Policies} 

\author{Michele Colledanchise, Giuseppe Cicala, Daniele E. Domenichelli, Lorenzo Natale, Armando Tacchella 
\thanks{Michele Colledanchise, Daniele E. Domenichelli, and 
  Lorenzo Natale are with Istituto Italiano di Tecnologia, via Morego 30, 16163 Genova. ({\tt\small name.surname@iit.it}) }
  \thanks{Giuseppe Cicala and Armando Tacchella are with Universit\`a degli Studi di Genova, DIBRIS ({\tt\small name.surname@unige.it})}%
\thanks{This work was carried out in the context of the SCOPE project, which has received funding from the European Union's Horizon 2020 research and innovation programme under grant agreement No 732410, in the form of financial support to third parties of the RobMoSys project.}
}

\maketitle
\thispagestyle{empty}
\pagestyle{empty}

\begin{abstract}
In this paper, we enable automated property verification of
deliberative components in robot control architectures. We focus on formalizing the execution context of Behavior Trees (BTs)
to provide a scalable, yet formally grounded, methodology to enable
runtime verification and prevent unexpected robot behaviors. To this end, we consider a message-passing
model that accommodates both synchronous and asynchronous composition
of parallel components, in which BTs and other
components execute and interact according to the communication patterns
commonly adopted in robotic software architectures. We
introduce a formal property specification language to encode
requirements and build runtime monitors. We performed a set of experiments, both on simulations and on the real robot, demonstrating
the feasibility of our approach in a realistic application and its integration in a typical robot software architecture. We also
provide an OS-level virtualization environment to reproduce the
experiments in the simulated scenario.

\end{abstract}

\section{Introduction}
\label{sec.introduction}
\input{introduction.tex}

\section{Scenario}
\label{sec:scenario}
\input{scenario.tex}


\section{{Representation of BTs}}
\label{sec:bg}
\input{background.tex}
\section{Model}
\label{sec:model}
\input{model.tex}

\newpage
\section{Runtime verification}
\label{sec:monitor}
\input{monitor.tex}

\section{Experimental results}
\label{sec:exp}
\input{experiments.tex}

\clearpage
\section{Concluding Remarks}
\label{sec:disc}

\input{discussion.tex}
%
\bibliographystyle{IEEEtran}
\bibliography{refs}
\end{document}

%% file: introduction.tex
Behavior trees (BTs) are a graphical model to specify reactive, fault-tolerant task executions. Behavior developers introduced them in the computer game industry. They are gaining popularity in robotics because they are highly flexible, reusable, and well suited to define deliberative elements in the model-based design of control architectures.
The use of BTs in robotics~\cite{BTBook, ghzouli2020behavior} spans from
manipulation~\cite{rovida2017extended,
  zhang2019ikbt} to task planning
\cite{neufeld2018hybrid,safronov2020task}, human-robot
interaction~\cite{
  axelsson2019modelling,ghadirzadeh2020human} to
learning~\cite{banerjee2018autonomous},
autonomous vehicles~\cite{sprague2018improving}, and system
analysis~\cite{biggar2020principled,de2020reconfigurable,ogren2020convergence}.
Moreover, BTs are used in the Boston Dynamics's Spot SDKs to model
the robot's mission\footnote{\url{https://www.bostondynamics.com/spot2_0}}, in the Navigation Stack and the Task Planning System of
ROS2~\cite{macenski2020marathon,PlanSys2}. 

As elements of control architectures, BTs must concur to  
satisfy system-wide requirements. These include both \emph{safety}, 
e.g., the robot must never put humans or its mission at
risk, and \emph{response}, e.g., the robot should perform its
intended tasks. Requirements  
can be formalized in some logical language so that
compliance of BTs and other elements can be
algorithmically assessed either off-line, e.g., with static
analysis~\cite{hecht1977flow,DBLP:conf/popl/CousotC77} and model   
checking~\cite{queille1982specification,clarke1986automatic}, 
or on-line, e.g., with runtime verification~\cite{leucker2009brief}.
The BT literature addresses the verification problem to some extent.
There are works providing formal 
semantics of BTs using either description
logic~\cite{klockner2013interfacing}, non-linear
algebra~\cite{ColledanchiseTRO17}, 
and pseudocodes~\cite{BTBook,giunchiglia2019conditional};
some authors proposed a framework to
synthesize BTs that satisfy specifications in 
Linear Temporal Logic (LTL)~\cite{colledanchise2017synthesis,
  tumova2014maximally}, and others use
LTL to specify the semantics of BT nodes and their
composition~\cite{biggar2020framework}. However, we argue
that these works do not address the problems of $(i)$ embedding BTs in
a context including other elements of the software architecture that
are required for the BT execution and $(ii)$ verifying that the stated
requirements are fulfilled in such context. 

We tackle these problems by providing a formalization of BTs that
considers the elements that they orchestrate, the services provided by
the robot at 
the functional level and the communication among all these elements,
and a property specification language to build and deploy monitors for
runtime verification. Our approach is consistently based on the
framework proposed by RobMoSys\footnote{\url{https://robmosys.eu/}}. In particular,
with reference to RobMoSys Composition Structures,
we categorize the BT as a \emph{task plot}, i.e., a sequence of
tasks required to achieve certain goals at runtime; the leaves of the
BT communicate with \emph{skills}, i.e., the coordination of
functional components made accessible to task-plots; finally, the
pieces of software that execute code at the functional layer
are \emph{components}. We argue that ours is  a formal, yet scalable,
approach that can grant an additional level of confidence when
deploying a robot in dynamic and unpredictable environments --- a
setting where BTs found large advantages over classical task execution
architectures.

More specifically, we contribute a model of BTs, skills,
  components and their interactions in terms of \emph{channel systems}
  composed by \emph{program
    graphs}~\cite{baier2008principles}. Intuitively, program graphs
  are extended finite state machines (FSMs) that provide an
  operational semantics for the execution of single BT nodes, skills
  and components. The overall system is assembled through
  \emph{communication actions} that formalize the parallel execution
  of the program graphs. We consider both synchronous actions, e.g.,
  to model communication among BT nodes, and asynchronous actions,
  e.g., to model communication (typically through the middleware) 
among skill and components, obtaining a
  \emph{globally-asynchronous locally-synchronous} model which matches
  well the actual implementation of the robot's control
  architecture. We present a human-readable logic language --- a
  subclass of Timed Propositional Temporal Logic
  (TPTL)~\cite{DBLP:journals/jacm/AlurH94} --- that enables us to
  translate requirements into formal properties about data exchanged
  through channels and extract monitors from them. {We
    chose TPTL over 
  plain Linear Temporal Logic (LTL) because TPTL can express real-time 
  properties like ``the robot must start heading towards the charging
  station at most thirty seconds after receiving a low-battery
  signal''. A dialect of TPTL, rather than the full logic, was chosen
  to simplify writing properties in a controlled natural language.}
  We also provide a formal model for the execution of monitors in
  parallel with other components. Finally, we validate our approach in
  a service robot scenario, both in simulation and with the R1
  humanoid robot~\cite{parmiggiani2017design}. To make our 
  simulated experiments reproducible, we provide the related software
  in an OS-level virtualization environment based on Docker.

%% file: scenario.tex
\begin{figure}[h!]
\centering
\includegraphics[width=\columnwidth]{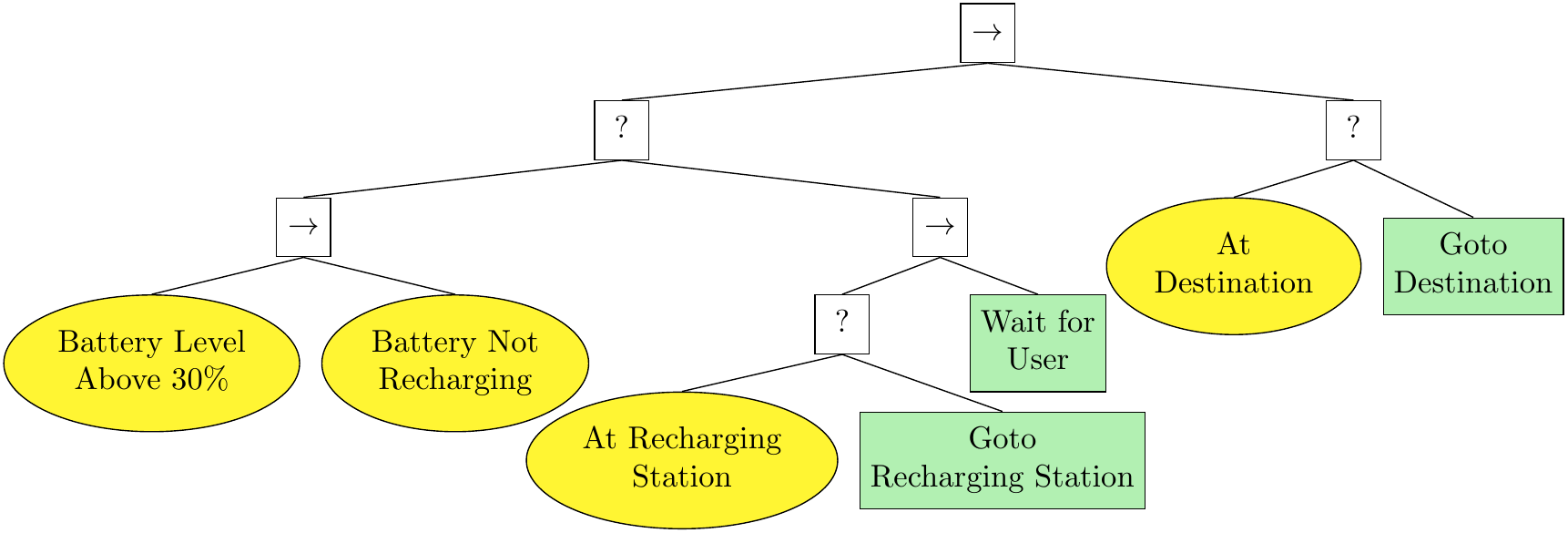}
\caption{BT for the validation scenario.}
\label{s1.fig.complexBT}
\end{figure}

A service robot must go to a predefined location. Whenever the battery goes
below $30$\% of its full capacity, the robot must stop and reach
the charging station, where it waits until the battery gets fully charged. Once the battery gets fully charged, the robot resumes the
previous navigation task. Figure~\ref{s1.fig.complexBT} shows the BT executed on the robot to accomplish the task.
As customary~\cite{BTBook}, we represent nodes and trees using a graphical 
syntax, where green rectangles indicate action
nodes, yellow ellipses indicate condition nodes, 
while ``$\rightarrow$'' and ``?'' indicate ``sequence'' and ``fallback'' nodes,
respectively. The execution of a BT
begins with the root receiving a ``tick'' signal that it propagates  
to its children. A node in the tree is executed if and only if it
receives tick signals. When the node no longer receives ticks,
its execution stops. Each child returns to the parent its status,
according to the specific logic it implements. In particular, 
when a sequence node receives ticks, it routes them to
its children from the first to the $N$-th. It returns \emph{failure}
or \emph{running} whenever a child returns so.
It returns \emph{success} whenever all the children return
success. The fallback also routes the tick signal to its
children from the first to the $N$-th. However, it returns \emph{success} or \emph{running}
whenever a child returns such status. It returns \emph{failure} whenever all its children
return \emph{failure}. The logic of the leaf nodes
is described below.

\paragraph*{Battery Level Above 30\% (Condition)}
Whenever this condition receives a tick, it sends a request to a
battery reader component that provides the battery level. The
condition returns \emph{success} if the battery level is above 30\% of its
full capacity. It returns \emph{failure} otherwise.  

\paragraph*{Battery Not Recharging} 
Whenever this condition receives a tick, it sends a request to a
battery reader component that provides the battery's charging status. The condition returns \emph{success} if the battery is not
recharging. It returns \emph{failure} otherwise. 

\paragraph*{At Location}
Whenever this condition receives a tick, it sends a request to a
localization component, which implements an Adaptive Monte Carlo
Localization~\cite{randazzo2018yarp}.  The condition returns \emph{success}
if the robot is at the prescribed location, and failure otherwise.

\paragraph*{Goto Location}
Whenever this action receives a tick, it sends a request to a
navigation component (in our implementation we use the YARP Navigation
Stack~\cite{randazzo2018yarp}), and then it waits for the outcome of
the navigation (destination reached or path not found). The action
returns \emph{failure} if the navigation component cannot find a
collision-free path to Location. It returns \emph{success} if the
robot reaches a destination. It returns running if the robot is
navigating towards the destination. Whenever this action receives an
halt, it sends a request to the navigation component to stop the
robot's mobile base. 

\paragraph*{Wait for User }
Whenever this condition receives a tick, it returns \emph{running}.

We impose some requirements on the execution of the BT in Figure~\ref{s1.fig.complexBT}. The first
one is that the battery level must never reach below 20\%.  More
precisely, the value read by the condition \emph{Battery Level above
  30\%} from the battery reader component must never be less than 20\%
of the actual battery capacity. This is because we assume that about
10\% of the battery will be sufficient to reach the recharging station
from any position.
The second requirement is that whenever the battery level reaches below 30\% of
its charge while the robot is going to its destination, the robot must
eventually go to the recharging station; as before, it is the value
read by the condition \emph{Battery Level above 30\%} from the battery
reader that we should consider to verify this requirement, and then
also the commands sent by the action \emph{Goto Recharging Station} to
the navigation components that we should check.

%% file: background.tex
To describe the formal semantics of BTs in their
surrounding context, we consider \emph{channel
  systems}~\cite{baier2008principles}, i.e., parallel systems where 
processes communicate via first-in-first-out buffers that may contain
messages. {A channel system formalizes a set of communicating components in which $(i)$ every component can be formalized as a finite state machine running concurrently with others and $(ii)$ communication happens asynchronously unless synchronization is forced over channels whose buffers have queues of length 0. The parallel execution of the components is assumed to be asynchronous unless explicit synchronization is provided.} 
Formally, let $Proc$ be a set of processes, and $Chan$ be a set of
\emph{channels} defined as
$$
Chan = \left\{ (p, q) \in Proc \times Proc \; | \; p \neq q\right\}.
$$
Let $Var$ be a set of variables over some domain, where $dom(x)$
denotes the domain of $x \in Var$, and let $dom((p,q))$
denote the domain of the messages transmitted over $(p,q)$.
We can define the set of \emph{communication actions} as 
\begin{align*}
  Comm = \{ & !(p,q,x), !(p,q,{<}m{>}), \\
            & ?(q,p,x), ?(q,p,{<}m{>}) \; | \\
            & (p,q) \in Chan,\; m \in dom((p,q)),\\ 
            & x \in Var \mbox{ with } dom(x) \supseteq dom((p,q)) \}.
\end{align*}
Actions denoted with a
``$!$'' are \emph{send actions}: $!(p,q,x)$ and
$!(p,q,{<}m{>})$ send the value of variable $x$ or a specific message
${<}m{>}$ from $p$ to $q$, respectively;
actions denoted with ``$?$'' are \emph{receive actions}:
with $?(q,p,x)$ and $?(q,p,{<}m{>})$ process $q$ receives
either the value to be assigned to a variable $x$ or a specific
message ${<}m{>}$ from $p$, respectively. 
A channel has a \emph{capacity} indicating the
maximum number of messages it can store; we denote with
$cap((p,q)) \in \mathds{N}$
the capacity of a channel $(p,q)$ with $cap((p,q)) \leq 1 $.
{Note that the special case $ cap((p,q)) = 0 $ is permitted. In this case, a communication corresponds to synchronous transmission. 
When $ cap((p,q)) > 0 $, there is a ``delay'' between the transmission and 
the receipt of a message, i.e., sending and
reading of the same message take place asynchronously.}
To model processes, we consider a set of variables $Var$,
where  $Eval(Var)$ denotes the set of \emph{evaluations} that assign
values to variables, and $Cond(Var)$ denotes a set of \emph{Boolean
  conditions} over the variables in  $Var$. For the sake of brevity,
we do not introduce conditions formally, but we just mention that they
are built using standard Boolean connectives and relational
operators over variables.
Given a set of channels $Chan$, communication actions $Comm$, and 
variables $Var$, a \emph{program graph} $PG$ over
$(Var, Chan)$ defined as
$$
PG = (Loc, Act, Effect, \hookrightarrow, Loc_0 , g_0)
$$
where:
\begin{itemize}
	\item $ Loc $ is a set of locations and $ Act $ is a set of actions,
	\item $  Effect : Act \times Eval(Var) \rightarrow Eval(Var) $ is the effect function,
	\item  $ \hookrightarrow \; \subseteq \; Loc \times (Cond(Var) \times (Act \cup Comm) \times Loc $ is the  transition relation,
	\item  $ Loc_0 \subseteq Loc $ is a set of initial locations,
	\item  $ g_0 \in Cond(Var) $ is the initial condition.
\end{itemize}
Given a set of  $n$ processes $Proc = \{ p_1,  \dots ,  p_n \} $ where
each $p_i$ is specified by a \emph{program graph} $PG_i$,  
a \emph{channel system} $CS = [PG_1 | \dots | PG_n]$ over $ (Var,
Chan) $ consists of the program graphs  $PG_i$ over $(Var_i , Chan)$ 
(for $  1 \leqslant i \leqslant n $) with $  Var = \bigcup_{1
  \leqslant i \leqslant n}  Var_i $. 

Intuitively, the transition relation $ \hookrightarrow $ of a program
graph over $(Var, Chan)$ consists of two types of conditional
transitions. The conditional transitions $ \ell
\xhookrightarrow{g:\alpha} \:  \ell' $ are labeled with guards and
actions: the condition $ g $ is the guard of the conditional
transition $ g:\alpha $.  
These conditional transitions can happen whenever the guard holds.
Alternatively, conditional transitions are labeled with
communication actions. This yields conditional transitions of type
$ \ell  \xhookrightarrow{g:!(p,q,\odot)} \:  \ell' $
with $\odot \in \{x, {<}m{>}\}$ for sending the value of a variable $x$
or a specific value ${<}m{>}$ along channel $ (p,q) $, 
or 
$ \ell  \xhookrightarrow{g:?(q,p,\odot)} \:  \ell' $
for receiving variables and values.
Based on the current variable evaluation, the capacity and the content
of the channel $(p,q)$, the guard is satisfied in the following cases:
\begin{itemize}
\item \emph{Handshaking}. When $ cap((p,q)) = 0 $, then process $p$
  can perform action $\ell_p  \xhookrightarrow{!(p,q,\odot)} \:
  \ell'_p $ only if  process $q$ offers the  complementary receive
  action $\ell_q  \xhookrightarrow{?(q,p,\odot)} \:  \ell'_q$;
\item \emph{Asynchronous message passing}. If $ cap((p,q)) = 1$, then
  process $p$ can perform the conditional transition
  $\ell_p  \xhookrightarrow{!(p,q,\odot)} \:  \ell'_p \quad$
  if and only if channel $(p,q)$ is empty.
  Accordingly, $q$ may perform the conditional transition
  $\ell_q  \xhookrightarrow{?(q,p,\odot)} \:  \ell'_q \quad$
  if and only if the channel $(p,q)$ is not empty; if $\odot = x$,
  then the message in the channel is extracted and assigned to $x$,
  whereas if $\odot = {<}m{>}$ the message is extracted as long as it
  is exactly ${<}m{>}$.
\end{itemize}
The execution of a channel system $CS = [PG_1 | \dots | PG_n] $ 
over $ (Chan, Var)$
is defined formally considering the equivalent transition system, 
but we skip the definition here since it is not necessary to
understand the rest of the paper --- details can be found
in the  literature~\cite{baier2008principles}. {One can see the execution of the channel system as the (a)syncrhonous parallel of all the state machines corresponding to the program graphs in the system. At each time, a state machine which can execute an action is chosen non-deterministically to perform that action. In case of synchronous communication actions, two or more state machines will perform transitions accordingly. In case of asynchronous communication actions, or an internal update, only a single machine will make a transition. This behavior abstracts over all possible scheduling techniques that provide some synchronization mechanism among processes.} 

%% file: model.tex
\begin{figure}[t!]
\centering
    \includegraphics[width=\columnwidth]{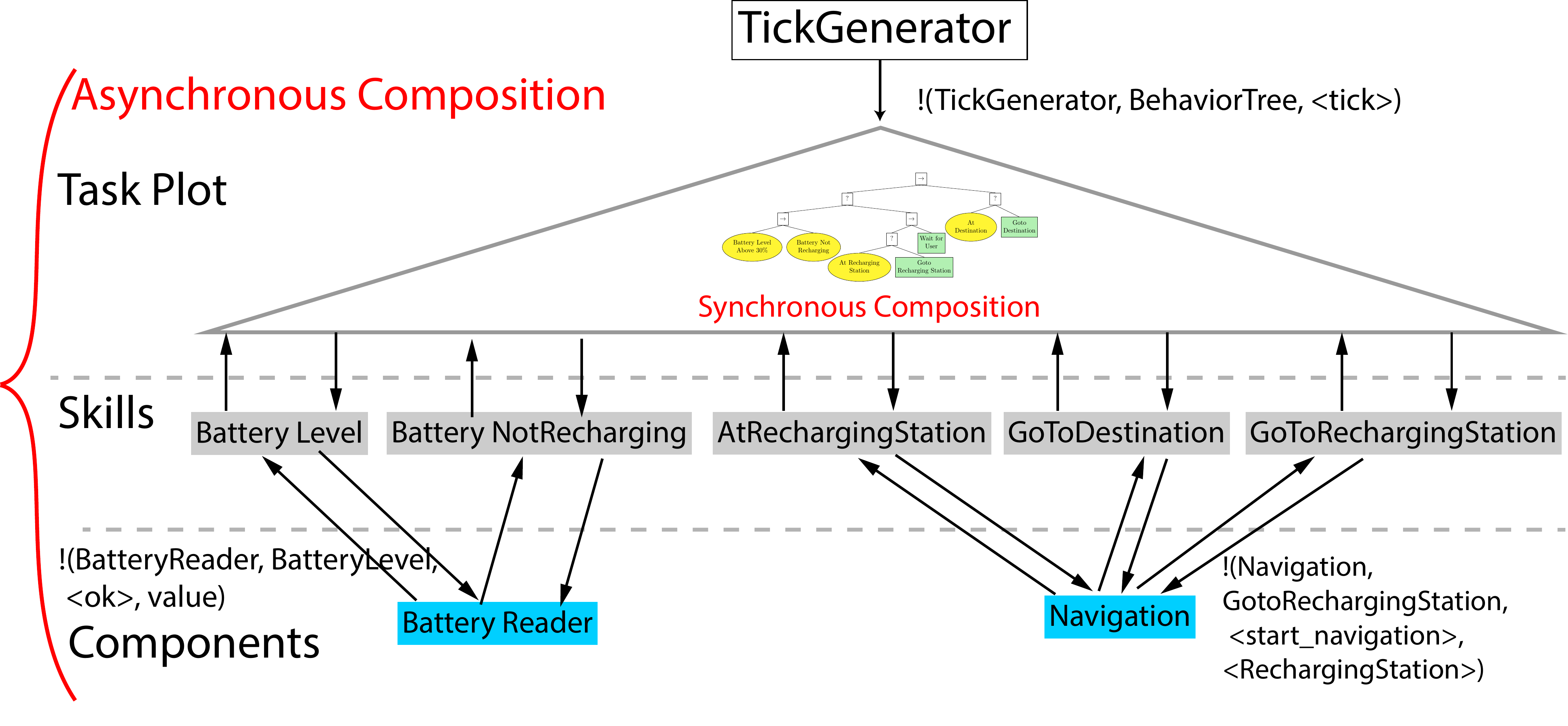}
  \caption{Graphical representation of
    the BT in Figure~\ref{s1.fig.complexBT} plus its surrounding
    context of skills and components. The triangle stands for the BT
    process, grey boxes represent skill processes, blue boxes
    represent component processes and arrows represent channels.} \label{fig:btcontext}
\end{figure}

In Figure~\ref{fig:btcontext} we show the BT of
Figure~\ref{s1.fig.complexBT} together with part of its surrounding
context. As we mentioned before, we structure all
the elements of our design according to the RobMoSys Composition
Structures, and the levels relevant for our design are
reported in Figure~\ref{fig:btcontext}. At the task plot level sits
the BT, receiving input from a \textsf{TickGenerator} process
whose task is simply to keep supplying tick signals to the
BT.\footnote{The formal notation we introduced in
  Section~\ref{sec:bg} is for closed systems, so we cannot
  model ticks coming from outside of the BT context.}
The communication action shown in the figure occurs each time the tick
generator sends a tick. The BT itself is a composition of processes,
one for each
 node. 
We assume that all the nodes exchange messages among them using handshaking,
i.e., the BT is a subsystem whose nodes interact in a synchronous
way. While we do not give the details of the internal semantics of the
nodes here, we wish to point out that this choice enables us to obtain
the (compositional) semantics of a BT as a \emph{deterministic}
transition system. In other words, a given BT is guaranteed to execute
in the same way given the same inputs, i.e., the tick signal and the
messages sent from the skills to the leaves.

In Figure~\ref{fig:btcontext} we depict five
out of the seven skills required by the BT (skill level) and two components 
interacting with the skills shown (component level). We assume that each skill and 
component is modeled either as a single process, or a composition of
processes. In the latter case, communication among such processes
must occur either through handshaking or shared variables, i.e.,
asynchronous communication is not allowed inside skills and
components, but it is allowed among components.  On the other hand, we assume that all the channels among BT, skills and components --- depicted as arrows in
Figure~\ref{fig:btcontext} --- have the capacity of one message at a time, i.e., the overall composition is asynchronous. For instance,
when requested by the skill \textsf{BatteryLevel}, the component
\textsf{BatteryReader} sends back the battery level in the message
\textsf{[$<$ok$>$, value]} where \textsf{value} represents the battery
level value acquired from the battery driver. 
Similarly, when the skill \textsf{GoToRechargingStation} is invoked
by the BT, it sends to the \textsf{Navigation} component a message
\textsf{[$<$start\_navigation$>$, $<$RechargingStations$>$]} in order
to reach the charging dock. 

In our model there is no direct communication
from the BT to the components at the functional level. Each skill related to a leaf node in the BT
coordinates one of several components to send commands to the robot and computes the return status. This choice is consistent with the RobMoSys Composition Structure diagram.\footnote{\url{https://robmosys.eu/wiki/modeling:composition-structures:start}}
Our choice of channels with capacity of a
single message at a time implies that no buffering occurs, and this
allows us to reproduce very closely concurrency schemes like Remote
Procedure Call (RPC). This  may look limiting considering
the wide adoption of publish-subscribe communication patterns, but 
it is consistent with the interfaces provided, e.g., by ROS services
and YARP modules with RPC interface, which are fairly typical in
applications of BTs.
In principle, the limitation of a single message
per channel can be lifted to model more complex communication
patterns, but this would also make verification harder.
Finally, we should note that asynchronous
composition introduces an element of non-determinism, i.e., scheduling
of processes is arbitrary under the assumption that only one
transition can be taken at any given time. However, when
multiple transition guards are true, the order in which they are taken
is arbitrary, and this reflects a level of abstraction with respect
to actual scheduling policies.


%% file: monitor.tex
The requirements defined in Section~\ref{sec:scenario} translate to
\emph{safety properties}, i.e., an event always/never occurs during
the execution of the BT, and to \emph{response properties}, i.e., it
must always be the case that if some event occurs during the execution
of the BT, then another one must eventually occur at a later time.
Formally, violations of safety properties can be found along with the
executions of finite length, while response properties can be  
violated only on executions of \emph{infinite}
length~\cite{baier2008principles}. 
Safety properties are thus monitorable, but response ones are not,
because the response could happen one step beyond the end of any
finite trace~\cite{leucker2009brief}.
To overcome this problem, we formalize properties through a
\emph{real-time} logic so that we can state, e.g., that the
robot starts moving to the recharging station \emph{within a specific
  delay} after seeing a low battery condition. 
In the remainder of this section, we 
introduce a real-time logic language and we show how to formalize
requirements and how to extract runtime monitors. 

\subsection{From requirements to formal properties}
\label{sub:encoding1}

The syntax of our property specification language
is inspired by the OTHELLO language~\cite{cimatti2013validation}, 
but we consider a semantic-based on timed state
sequences~\cite{DBLP:journals/jacm/AlurH94} instead of 
one based on hybrid traces~\cite{cimatti2013validation}. Our
language is a fragment of Timed Propositional Temporal Logic
(TPTL)~\cite{DBLP:journals/jacm/AlurH94} and has the grammar shown in 
Figure~\ref{fig:othello}. We consider the standard abbreviations for
Boolean connectives and constants, 
e.g., we write ``\ote{false}'' for ``\ote{not} \ote{true}'', ``$\alpha$
\ote{implies} $\beta$'' for ``\ote{not} $\alpha$ \ote{or} $\beta$'';
similarly, for temporal connectives we write ``\ote{eventually}
$\alpha$'' for ``\ote{true} \ote{until} $\alpha$'' and ``\ote{always}
$\alpha$'' for ``\ote{not} \ote{eventually} \ote{not} $\alpha$''.  
The main feature of SCOPE is that atoms are
always related to channel \emph{events}, i.e., Boolean conditions that
become true when specific values are transmitted over channels. 
We consider two kinds of events: \emph{untimed events} do not have a
time delay associated with them, e.g., the atom
$( \textsf{BatteryReader}, \textsf{BatteryLevel},
  m[1] = <\mbox{ok}>  \mbox{ \ote{implies} } m[2] > 20\; )$ 
is either or true or false at a specific point in time depending on
the contents of the channel from \textsf{BatteryReader} to
\textsf{BatteryLevel}; on the other hand, \emph{timed events} have a
time window associated with them, e.g., the atom
$\ote{time\_until} ( \textsf{BatteryReader}, \textsf{BatteryLevel},
                     m[1] = <\mbox{ok}> \mbox{ \ote{and} }
                      m[2] < 30 ) < 10$
is either true or false depending on the contents of the channel
between \textsf{BatteryReader} and \textsf{BatteryLevel}
within 10 time steps from the current point in time, i.e., 
the channel  must contain the message
$[<\mbox{ok}>, v]$ with $v < 30$ at least once
within 10 time steps.

\begin{figure}
\begin{center}
\begin{tabular}{lcl}
  \onte{constraint} & := & \onte{atom} $\mid$ \\
  &    & \ote{not} \onte{constraint} $\mid$ \\
  &    & \onte{constraint} \ote{or} \onte{constraint} $\mid$ \\
  &    & \onte{constraint} \ote{and} \onte{constraint} $\mid$ \\
  &    & \ote{next} \onte{constraint} $\mid$ \\
  &    & \onte{constraint} \ote{until} \onte{constraint} \\
  &    & \ote{(} \onte{constraint} \ote{)}; \\  
  \onte{atom} & := & \ote{true} $\mid$ \onte{event} $\mid$ \\
  &    & \ote{time\_until} \ote{(} \onte{event} \ote{)} \onte{relop}
  \onte{const}; \\ 
  \onte{event} & := & \ote{(} \onte{source} \ote{,} \onte{dest} \ote{,} \onte{condition} \ote{)} \\ 
  \onte{condition} & := & \onte{message} \onte{relop} \onte{const} \\
  &    & \ote{not} \onte{condition} \\
  &    & \onte{condition} \ote{or} \onte{condition} \\
  &    & \onte{condition} \ote{and} \onte{condition} \\ 
\end{tabular}
\end{center}
\caption{\label{fig:othello} SCOPE language grammar: \onte{source} and
\emph{dest} are members of a given set of processes \emph{Proc};
\onte{const} is a constant of the appropriate domain ---
\onte{const} $\in \mathbb{N}$ for \onte{time\_until} constraints
and the same of \onte{message} in \onte{condition};
\onte{relop} is a relational operator in the set $\{ <, >, \leq, \geq,
=, \neq\}$. }
\end{figure}

We give here a brief and mostly informal account about the
semantics of SCOPE, leaving the details
to~\cite{DBLP:conf/memocode/DokhanchiHTF16} from which we borrow
the notation and the definitions.
The meaning of SCOPE constraints is given in terms of timed state
sequences generated by a channel system like the one described in
Section~\ref{sec:model} and shown in Figure~\ref{fig:btcontext}.
The component \textsf{TickGenerator} provides (discrete) \emph{time
  sequences} $\tau = \tau_0 \tau_1 \tau_2 .. $, i.e., sequences of  
tick identifiers $\tau_i \in \mathbb{N}$ and $i \in
\mathbb{N}$. The first tick generated will have identifier 0, then 1, 2,
and so on. We assume 
that time sequences are \emph{initialized}, i.e., $\tau_0 = 0$,
\emph{monotonic}, i.e., $\tau_i \leq \tau_{i+1}$ for all $i \in
\mathbb{N}$  and \emph{progressive}, which means that for all $t \in
\mathbb{N}$ there is some $i \in \mathbb{N}$ such that $\tau_i >
t$. Let $Chan$ represent the set of channels connecting the tick
generator, the BT, the skills and the components, and $Eval(Chan)$
be the set of evaluations on such channels. The execution of the
overall system provides \emph{state sequences} $\sigma = \sigma_0
\sigma_1 \sigma_2 \ldots$ where $\sigma_i \in Eval(Chan)$ and
$i \in \mathbb{N}$. 
A \emph{timed state sequence} (\emph{TSS}) $\rho =
(\sigma, \tau)$ is a pair consisting of a state sequence $\sigma$ and
a time sequence $\tau$ where $\rho_0 \rho_1 \rho_2 \ldots = (\sigma_0,
\tau_0) (\sigma_1, \tau_1) (\sigma_2, \tau_2) \ldots$. Intuitively, a
TSS pairs a tick identifier with a channel evaluation being consistent
with that identifier. Note that the same tick identifier might occur
several times in a TSS because a single tick usually implies a
number of changes in the overall channel evaluation as signals are
propagated from the BT to skills and to components \emph{without} new
ticks being generated. 

We write
$\rho \models \varphi$ to denote that the SCOPE constraint $\varphi$ is
\emph{sastified} by the TSS $\rho$. This relationship is formally
described for TPTL in~\cite{DBLP:conf/memocode/DokhanchiHTF16} and
thus holds also for our fragment SCOPE. Intuitively, Boolean formulas are 
satisfied at any point $i \in \mathbb{N}$ of a TSS if the evaluation
$\sigma_i$ makes the formula true, and temporal connectives have the
usual meaning: \ote{next} $\alpha$ is satisfied at a point in time if
$\alpha$ is satisfied at the next one; $\alpha$ \ote{until} $\beta$ is
satisfied if $\alpha$ holds until $\beta$ eventually does. Timed
events are handled as expected. For instance, 
\ote{time\_until} ( $\epsilon$ ) $<$ $\theta$ is satisfied at a
point $i$ in time if event $\epsilon$ occurs within step $i + \theta$ 
(excluded), whereas \ote{time\_until} ( $\epsilon$ ) $=$ $\theta$ 
is satisfied at a point $i$ in time if event $\epsilon$ occurs exactly
at step $i + \theta$.
Given a channel system $CS = [PG_1 | \dots | PG_n]$ 
an \emph{execution} of $CS$ is  any TSS $\rho$ consistent with the process composition, assuming that at least one process provides
the signal for the time trace --- in our case, it is \textsf{TickGenerator}. 
Given a SCOPE constraint $\varphi$ whose events occur over the channels
of $C$, we write $CS \models \varphi$ when $CS$ \emph{satisfies} the
constraint $\varphi$, i.e., for all the executions $\rho$ of $CS$ we
have that $\rho \models \varphi$. The safety and response requirements on the 
scenario described in Section~\ref{sec:scenario} become the SCOPE properties
depicted in Figure~\ref{fig:property}, where $\theta$ is some user-specified constant threshold.

\begin{figure}[t!]
\begin{equation*}
\small
  \begin{array}{l@{}l@{}l@{}l}
    \phi_1 = & \; \ote{always}  & \multicolumn{2}{l}{(\textsf{BatteryReader}, \textsf{BatteryLevel},}\\
    &               & \multicolumn{2}{l}{m[1] = <\!\mbox{ok}\!> \mbox{ \ote{implies} } m[2] >= 20)} \\ \\
    \phi_2 = & \; \ote{always} ( & \multicolumn{2}{l}{(\textsf{Navigation}, \textsf{GoToDestination},} \\
             &                & \multicolumn{2}{l}{m[1] = <\!\mbox{ok}\!> \mbox{ \ote{and} } m[2] = <\!\mbox{running}\!>)\;\ote{and}} \\
             &                & \multicolumn{2}{l}{(\textsf{BatteryReader}, \textsf{BatteryLevel},} \\
             &                & \multicolumn{2}{l}{m[1] = <\!\mbox{ok}\!> \mbox{ \ote{and} } m[2] <= 30)} \\
             &                & \multicolumn{2}{l}{\ote{implies time\_until ( }} \\
             &                &               & \quad \textsf{GotoRechargingStation}, \textsf{Navigation}, \\
             &                &               & \quad m[1] = <\!\mbox{start\_navigation}\!> \mbox{ \ote{and} } \\
             &                &               & \quad m[2] = <\!\mbox{RechargingStation}\!>) < \theta
  \end{array}
\end{equation*}
\vspace*{-1em}
\caption{\label{fig:property} Requirements as SCOPE properties.}
\end{figure}


\subsection{From formal properties to runtime monitors}
\label{sub:encoding2}

\begin{figure}
  \begin{subfigure}[t]{0.4\columnwidth}
    \scalebox{0.17}{\includegraphics{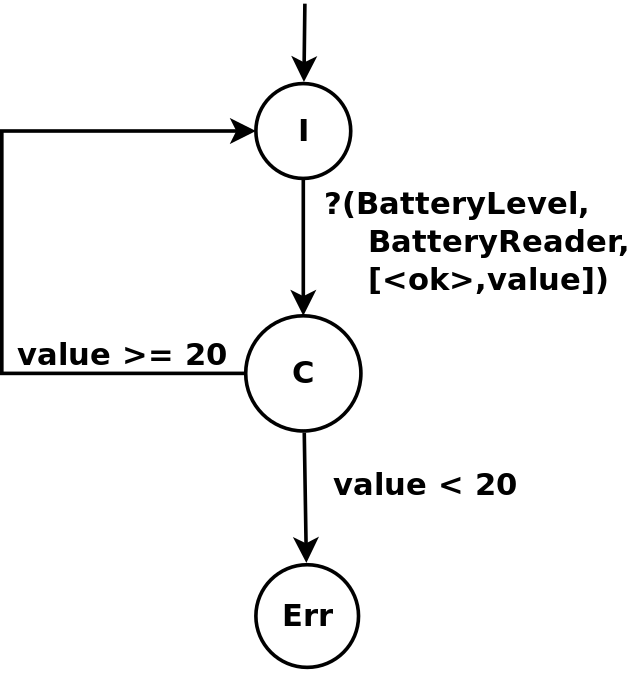}}
    \caption{\label{subfig:safety}\ Safety}
  \end{subfigure}
  \begin{subfigure}[t]{0.58\columnwidth}
    \scalebox{0.17}{\includegraphics{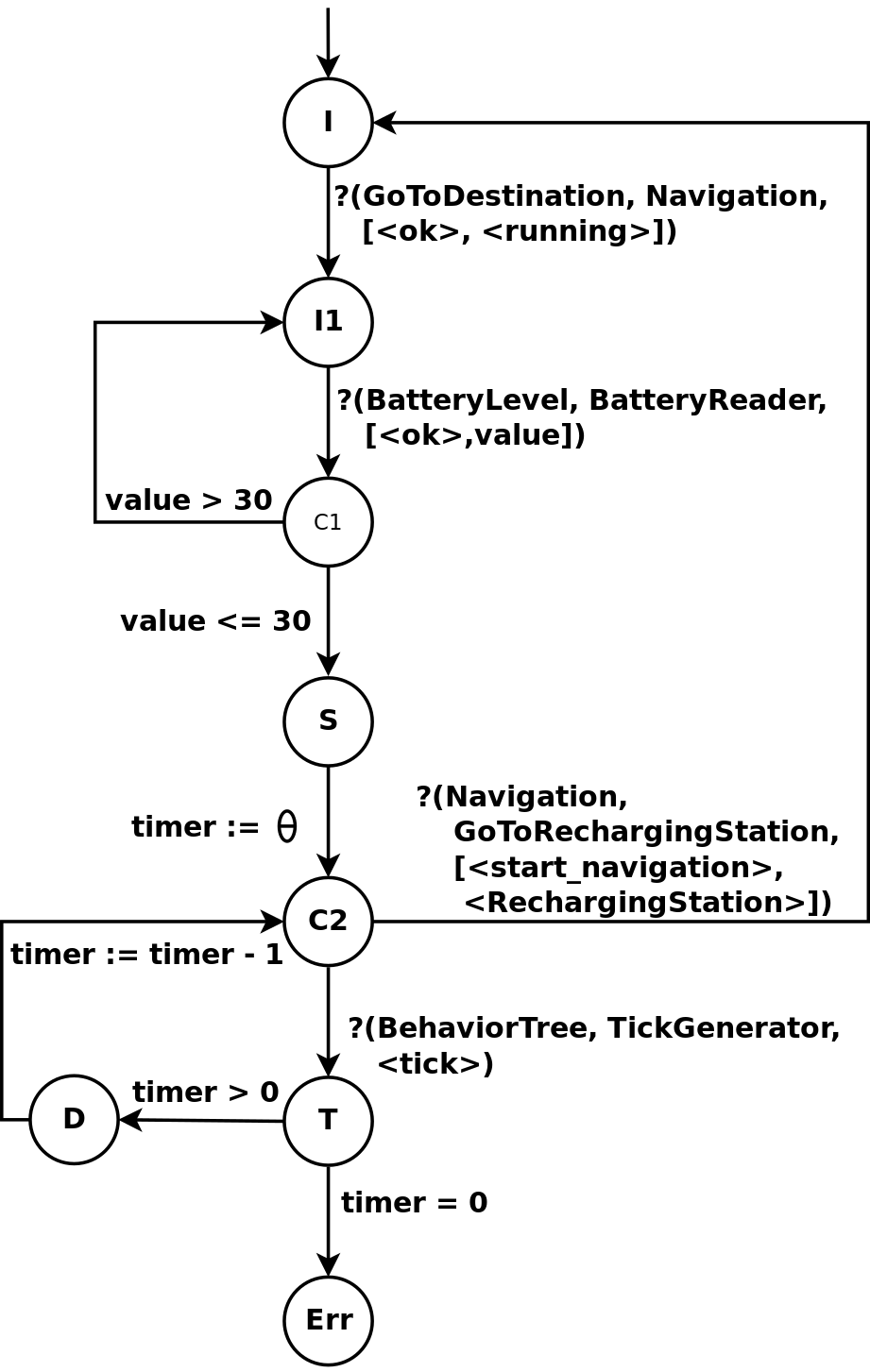}}
    \caption{\label{subfig:response}\ Response}
  \end{subfigure}
  \caption{\label{fig:monitor} Runtime monitors for safety and
    response properties ($\phi_1$ and $\phi_2$ respectively).}
\end{figure}

The literature provides a general monitoring
algorithm for TPTL properties~\cite{DBLP:conf/memocode/DokhanchiHTF16}. We could resort to
that procedure in order to obtain monitors from properties
written in SCOPE. However, since SCOPE is a fragment of TPTL and
we are interested in properties having a specific structure, we 
are able to obtain even more compact and efficient monitors. In
Figure~\ref{fig:monitor} we show the monitors (specified as program 
graphs) extracted from the properties stated above:
Figure~\ref{subfig:safety} shows the monitor for the safety property 
$\phi_1$ and Figure~\ref{subfig:response} shows the monitor for the
response property $\phi_2$. Looking at Figure~\ref{subfig:safety}, we can
observe that the safety monitor makes a transition from the initial
location \textsf{I} whenever the value of the battery charge is
transmitted from the component \textsf{BatteryReader} to the skill 
\textsf{BatteryLevel}. Intuitively, the monitor ``sniffs'' from the channel
the value transmitted checking whether it is greater than
or equal to 20, or less than 20: in the first case, it returns to the
initial location, waiting for new messages; in the second case, it enters
the location \textsf{Err} to signal that the property was violated in
the current execution. In Figure~\ref{subfig:response}, we can observe
that the response monitor makes a transition to the state
\textsf{I1} whenever the navigation component is answering \btrun{} to
the \textsf{GoToDestination} skill, meaning that the robot is
making its way to the destination. When the value of the battery
charge is transmitted, the monitor moves to state \textsf{C1}: if the
value is above 30, the monitor goes back to state \textsf{I1};
otherwise, it goes to state \textsf{S}, where a further transition sets
the variable \textsf{timer} to $\theta$ and moves to state
\textsf{C2}. Here, two things may happen: either $(i)$ the 
monitor detects a message on the channel from the skill
\textsf{GotoDestination} to the component \textsf{Navigation}
commanding the latter to go to the recharging station or $(ii)$ the
monitor detects a tick sent to the behavior tree. In the former case,
the monitor ``resets'' to the initial state \textsf{I}, as the
response property was fulfilled within $\theta$ ticks sent to the
BT. In the latter case, it checks whether \textsf{timer} reached 0, in 
which case it enters an error state, while if \textsf{timer} is still
greater than 0 it decrements \textsf{timer} and goes back in state
\textsf{C2} to wait for messages. While the program graphs described
in Figure~\ref{fig:monitor} are specific to properties $\phi_1$ and
$\phi_2$, it is easy to see how the construction can generalize to
safety and response properties having the same structure.

The last element that remains to be added to our description is
the interaction of monitors with the other processes in the channel
system $CS$ of Figure~\ref{fig:btcontext}. Let 
$CS' = [PG_1 | \dots | PG_n | PG_r]$ over $(Chan, Var)$ be the channel 
system with added monitor $PG_r$. We introduce
the concept of \emph{monitored transition} to allow monitors to make
transitions based on channel values without interfering with other
processes. For instance, let us consider monitoring asynchronous
message passing for channel $(p,q) \in Chan$, when process $q$
receives a value $ v \in dom((p,q)) $ from process $p$ and assigns it
to variable $ x \in Var $  with  $dom(x) \supseteq dom((p,q))$, while
monitor $r$ assigns it to variable $y \in Var$ with $dom(y) =
dom(x)$. Given $\eta \in Eval(Var)$ and $\xi \in Eval(Chan)$ as
current evaluations of variables and channels, respectively, the rule
for such a transition is:  
\newline
\begin{equation}
\small
  \frac{\ell_i  \xhookrightarrow{g:?(q,p,x) } \:  \ell'_i \; \wedge \; \eta \vDash g 
    \; \wedge \ell_r  \xhookrightarrow{g:?(q,p,x) } \: \ell'_r \; \wedge \; \xi((p,q)) = v}
       {\langle \ell_1, \ldots,  \ell_i, \ldots, \ell_n, \ell_r, \eta,\xi \rangle \xrightarrow{\tau}
					\langle \ell_1, \ldots,
                                        \ell'_i, \ldots, \ell_n, \ell'_r\eta',\xi' \rangle}
\end{equation}

where $ \eta' = \eta[x \coloneqq v; y \coloneqq v] $ and $ \xi' =
\xi[(p,q) \coloneqq \varepsilon] $. Notice that the
evaluation of \emph{both} $x$ and $y$ is changed, and the
location change occurs in the processes as well as in the monitor.
{Intuitively, the rule ``forces'' a synchronous transition between the component(s) to be monitored and the corresponding monitor, to make sure that all relevant signals are processed by the monitor at the same time in which they are transmitted. In practice, for instance in publish/subscribe middleware, this action can be simply implemented by having the monitor subscribe to all the topics of interest; clearly, we expect that only published topics will be subject to monitoring.}
All the other rules for transitions involving an exchange of data,
including those for handshaking, can be modified in a similar way to
take into account monitors. If multiple monitors are sniffing a
specific channel, then the transition should include all of them at
once, i.e., the update of monitors sniffing the same channel is synchronized.

%% file: experiments.tex
In this section, we present the experimental results. We made available the video\footnote{\url{https://youtu.be/ipLpRIh7Sp4}} of the experiments and the code in a pre-installed OS-level virtualization environment\footnote{\url{github.com/SCOPE-ROBMOSYS/IROS2021-experiments}} to reproduce them.

\paragraph*{Toolchain Overview}
We implemented the BT using the \emph{BehaviorTree.CPP} engine\footnote{\url{https://github.com/BehaviorTree/BehaviorTree.CPP}} and Finite State Machines (FSMs) for the skills and the runtime monitors using the \emph{Qt SCXML} engine.\footnote{\url{https://doc.qt.io/qt-5/qtscxml-overview.html}}
 We defined the communication between BT and each skill FSM, and between a skill FSM and the components it orchestrates using an Interface Definition Language handled via the YARP middleware~\cite{metta2006yarp}. We use a feature of the YARP middleware called portmonitor~\cite{paikan2014data} to transparently intercept the communication between the BT and the skills and between the skills and the components. The portmonitor then propagates the messages to the FSM of the corresponding runtime monitor.

\paragraph*{Real Robot}
We tested our approach on the IIT R1~\cite{parmiggiani2017design}
robot. Concerning the navigation capabilities, the robot's wheeled
base is equipped with two laser scanners, one at the front and one at
the back. We employ an Adaptive Monte Carlo Localization system to
localize the robot and an A$^*$-based algorithm to compute the path to
the destination~\cite{randazzo2018yarp}. The algorithms rely on the
odometry and the laser scan inputs.  

\paragraph*{Simulation Environment}
The robot is represented in the map as a circle and an orientation.
Starting from the center of the robot's representation, we 
compute the input of the laser scanner by casting radially polarized
beams, and we measure the collision on a point of the map labeled as
obstacles. We do not model sensor noise. We assume the absence
of uncertainty on the robot's initial position and we do not model the
disturbance on the robot movements. We employ the same localization
and navigation system of the real IIT R1. Notice that the
simulated robot has the same software interface of IIT R1 when
it comes to sensors and actuators. Therefore, it captures all
the complexity of the real system relevant to our work. 
Moreover, the runtime monitor mechanisms will detect
possible misbehaviors of the simulated model caused
by wrong assumptions about the model itself.  
\begin{experiment}[Runtime monitor for a safety property]
\label{experiment.battery}
This experiment, executed in the simulated environment, shows an execution example of the runtime monitor on the battery level on the scenario described in Section~\ref{sec:scenario}. The monitor implemented is the one depicted in Figure~\ref{subfig:safety}.
In particular, the monitor verifies that the property $\phi_1$ holds, i.e. \say{The battery level must never reach below 20\%}  (see Section~\ref{sec:monitor}).
 To impose a property's violation,  we inject a fault in the system by manually changing the level of the battery. 
Figure~\ref{exp1.fig} shows the execution steps, of both the scenario and the monitor, of this experiment. 
Initially, the battery level is above 30\% (Figure~\ref{exp1.fig.scenario1}). The BT sends ticks to the condition \say{Battery Level Above 30\%} and the condition sends a request to the corresponding skill. Then, the skill sends a request to the battery component to request the battery level. The portmonitor detects the request and sends the corresponding message to the runtime monitor, which goes to the state \emph{get} (Figure~\ref{exp1.fig.exec1}). When the component replies to the skill, the portmonitor propagates the reply to the runtime monitor. While the battery value is above 20\% the monitor moves back to the state \emph{idle} (Figure~\ref{exp1.fig.exec2}).
When we manually impose the battery level to be 10\% (Figure~\ref{exp1.fig.scenario2}) the monitor detects the property violation, thus it goes to the state \emph{failure} (Figure~\ref{exp1.fig.exec3}).

\begin{figure}[t]
\centering
\begin{subfigure}[t]{0.49\columnwidth}
\includegraphics[width=\columnwidth]{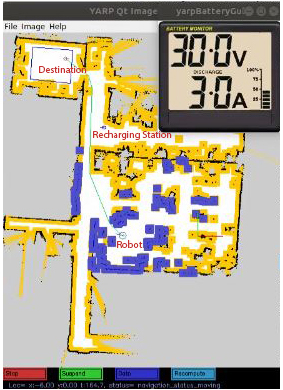}
\caption{Battery level above 30\%. The robot is reaching the destination. The safety property $\phi_1$ is satisfied.}
\label{exp1.fig.scenario1}
\end{subfigure}
\begin{subfigure}[t]{0.49\columnwidth}
\includegraphics[width=\columnwidth]{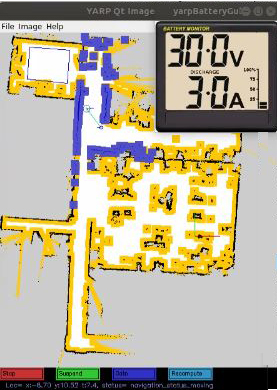}
\caption{Battery level below 30\%. The robot is reaching the recharging station. The safety property $\phi_1$ is violated.}
\label{exp1.fig.scenario2}
\end{subfigure}

\vspace*{1em}
\begin{subfigure}[t]{0.32\columnwidth}
\includegraphics[width=\columnwidth]{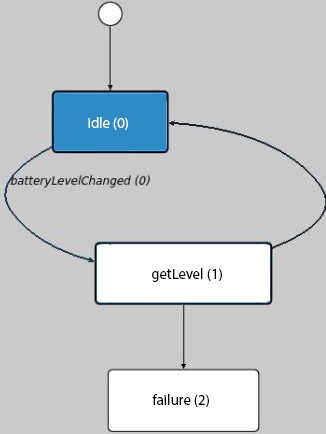}
\caption{The monitor at state Idle; waiting for the skill to send a request to the component.}
\label{exp1.fig.exec1}
\end{subfigure}
\begin{subfigure}[t]{0.32\columnwidth}
\includegraphics[width=\columnwidth]{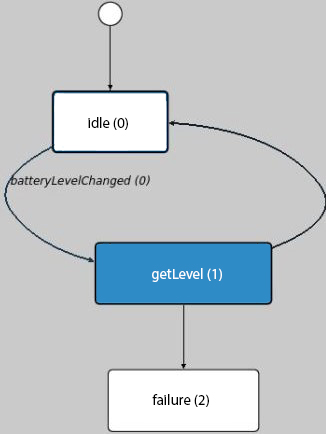}
\caption{The monitor at state Get; waiting for the component to respond.}
\label{exp1.fig.exec2}
\end{subfigure}
\begin{subfigure}[t]{0.32\columnwidth}
\includegraphics[width=\columnwidth]{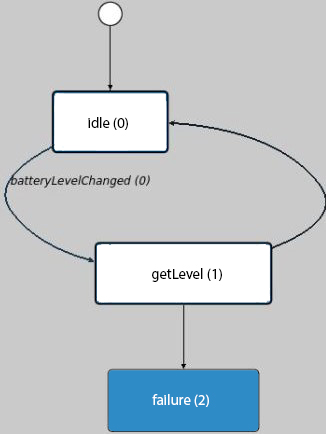}
\caption{The monitor at state Failure. Property violated.}
\label{exp1.fig.exec3}
\end{subfigure}
\caption{Execution steps and runtime monitors of Experiment~\ref{experiment.battery}. In the scenario, the destination is the room in the top left. The charging station is the small
circle on the way to the destination. The green curve is the computed path. The blue pixels represent
objects detected by the laser scanner.}
\label{exp1.fig}
\end{figure}

\end{experiment}

\begin{experiment}[Runtime monitor for a responsive property]
This experiment, executed in the simulated environment, shows an
execution example on the runtime monitor on the battery recharging
behavior. The monitor implemented is the one depicted in
Figure~\ref{subfig:response}.  In particular, the monitor verifies
that the property $\phi_2$ holds, i.e. \say{Whenever the battery level
  reaches below 30\% of its charge, the robot must eventually go to
  the recharging station} (see Section~\ref{sec:monitor}).
 To impose a property's violation, we introduce a bug in the FSM of
 the skill \say{Battery Level Above 30\%} such that it returns success
 while the battery level is above 20\%.  The execution is similar to
 the one of Experiment~\ref{experiment.battery}. The runtime monitor
 goes to an error state because the battery gets below 30\% during the
 robot navigation and before it gets to the charging station.
\end{experiment}

 \begin{experiment}[Monitor on a real robot]
\label{experiment.robot}
This experiment shows the execution with a safety property
violation in a real robot scenario.  In particular, the monitor
verifies that the property $\phi_1$ holds, i.e. \say{The battery level
  must never reach below 20\%} (see Section~\ref{sec:monitor}).  To
impose a property's violation, a malicious user tampers with the robot
control system, moving it with a joypad that overrides the navigation
commands, impeding the robot from reaching the charging station.

Figure~\ref{s2.fig.exec} shows the execution steps of this scenario,
while Figure~\ref{s2.fig.mon} shows the FSM of the runtime monitor.
When the battery level gets below 30\%, the robot moves towards the
charging station (Figure~\ref{s2.fig.exec1}). While the robot moves
towards the charging station, a malicious user moves the robot
elsewhere (Figure~\ref{s2.fig.exec2}). Then the robot resumes the
navigation (Figure~\ref{s2.fig.exec3}), and the malicious user moves
the robot elsewhere again. After several inputs from the malicious
user, the battery level gets below 20\% (Figure~\ref{s2.fig.exec4})
and the monitor detects the violation of the property
(Figure~\ref{s2.fig.mon2}).

\begin{figure}[h!]
\centering
\begin{subfigure}[t]{0.45\columnwidth}
\includegraphics[width=\columnwidth]{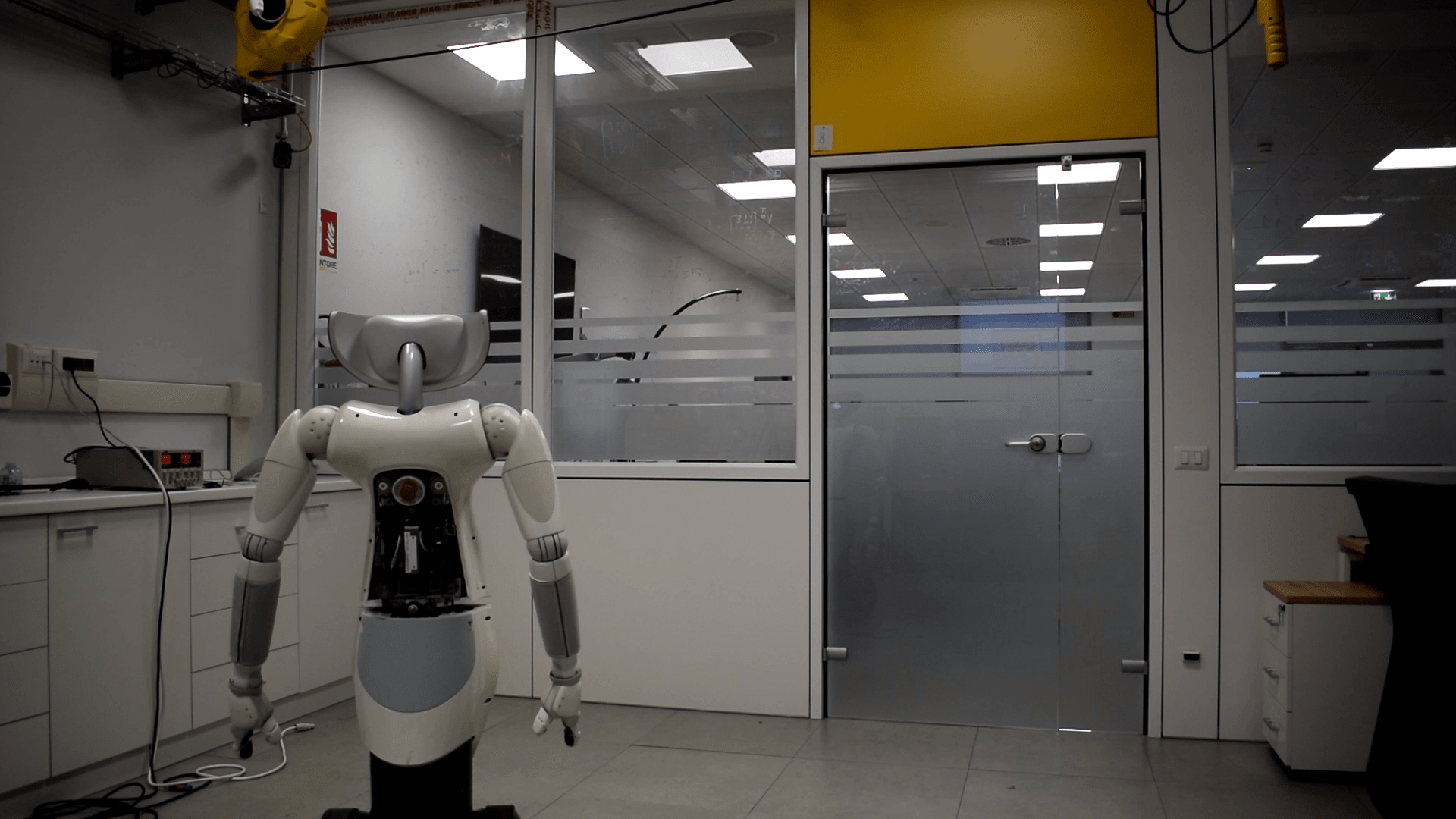}
\caption{The robot navigates to the charging station. }
\label{s2.fig.exec1}
\end{subfigure}
\begin{subfigure}[t]{0.45\columnwidth}
\includegraphics[width=\columnwidth]{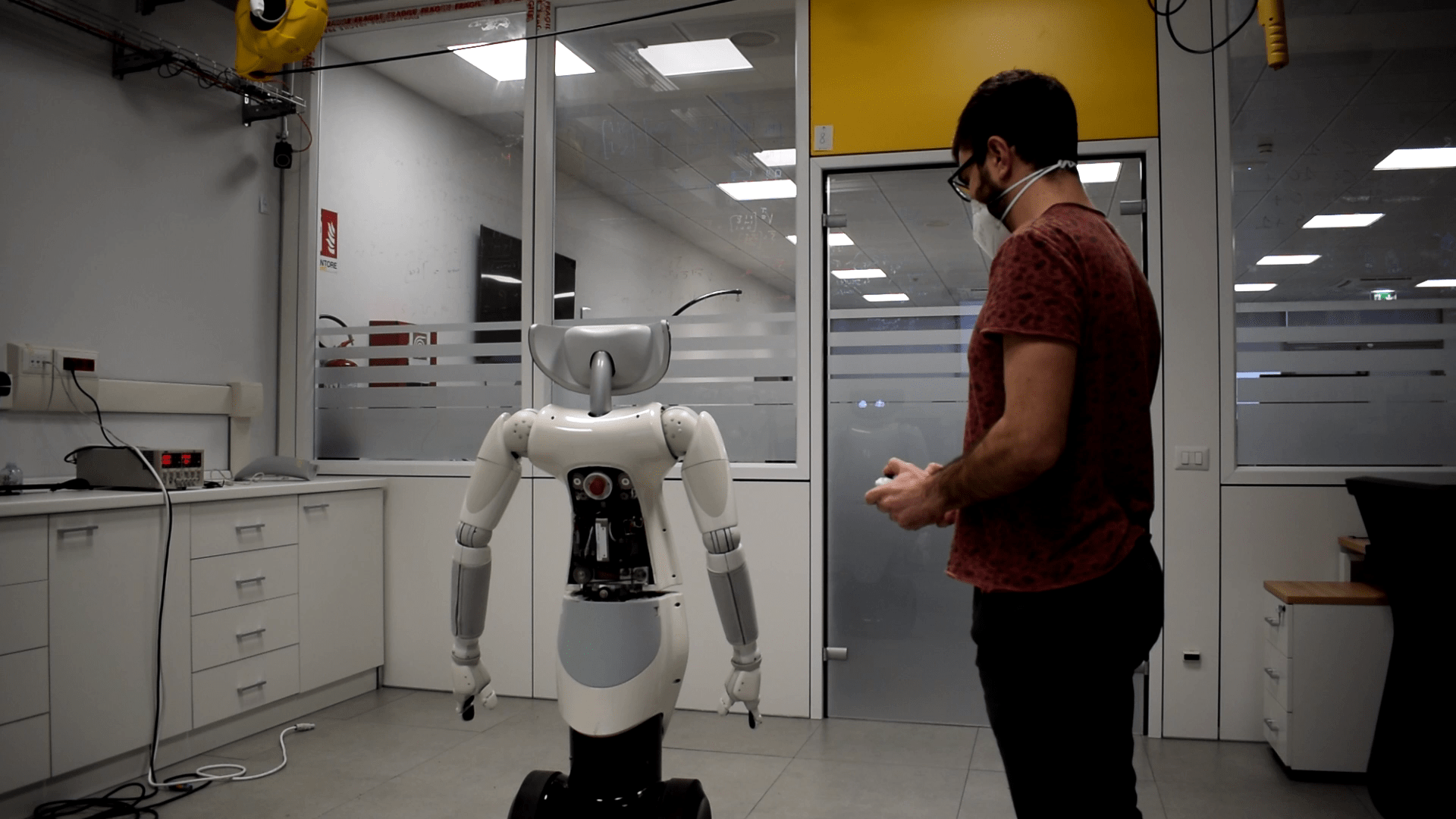}
\caption{A malicious user moves the robot elsewhere.}
\label{s2.fig.exec2}
\end{subfigure}

\vspace*{0.5em}
\begin{subfigure}[t]{0.45\columnwidth}
\includegraphics[width=\columnwidth]{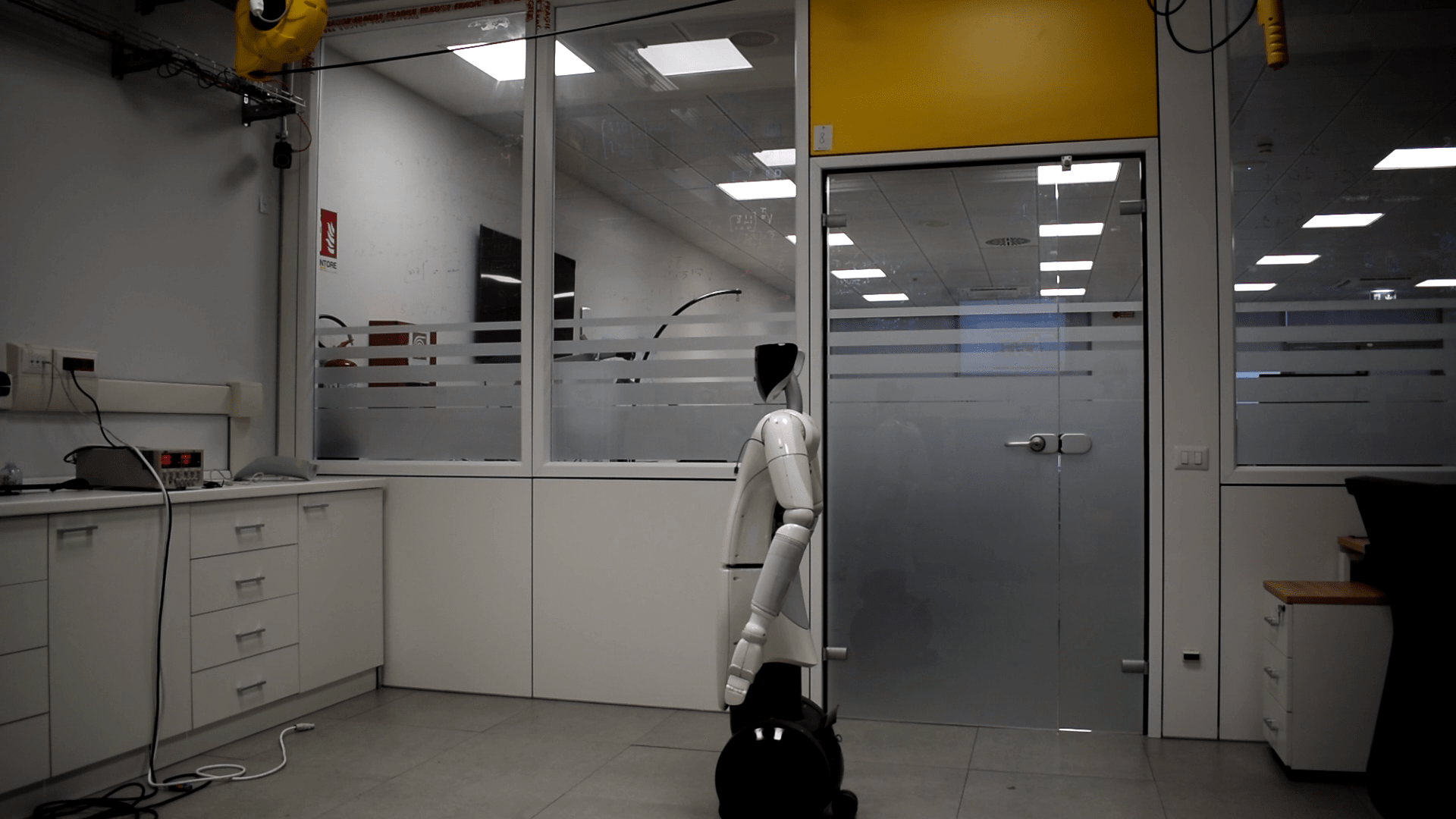}
\caption{The robot resumes its navigation to the charging station.}
\label{s2.fig.exec3}
\end{subfigure}
\begin{subfigure}[t]{0.45\columnwidth}
\includegraphics[width=\columnwidth]{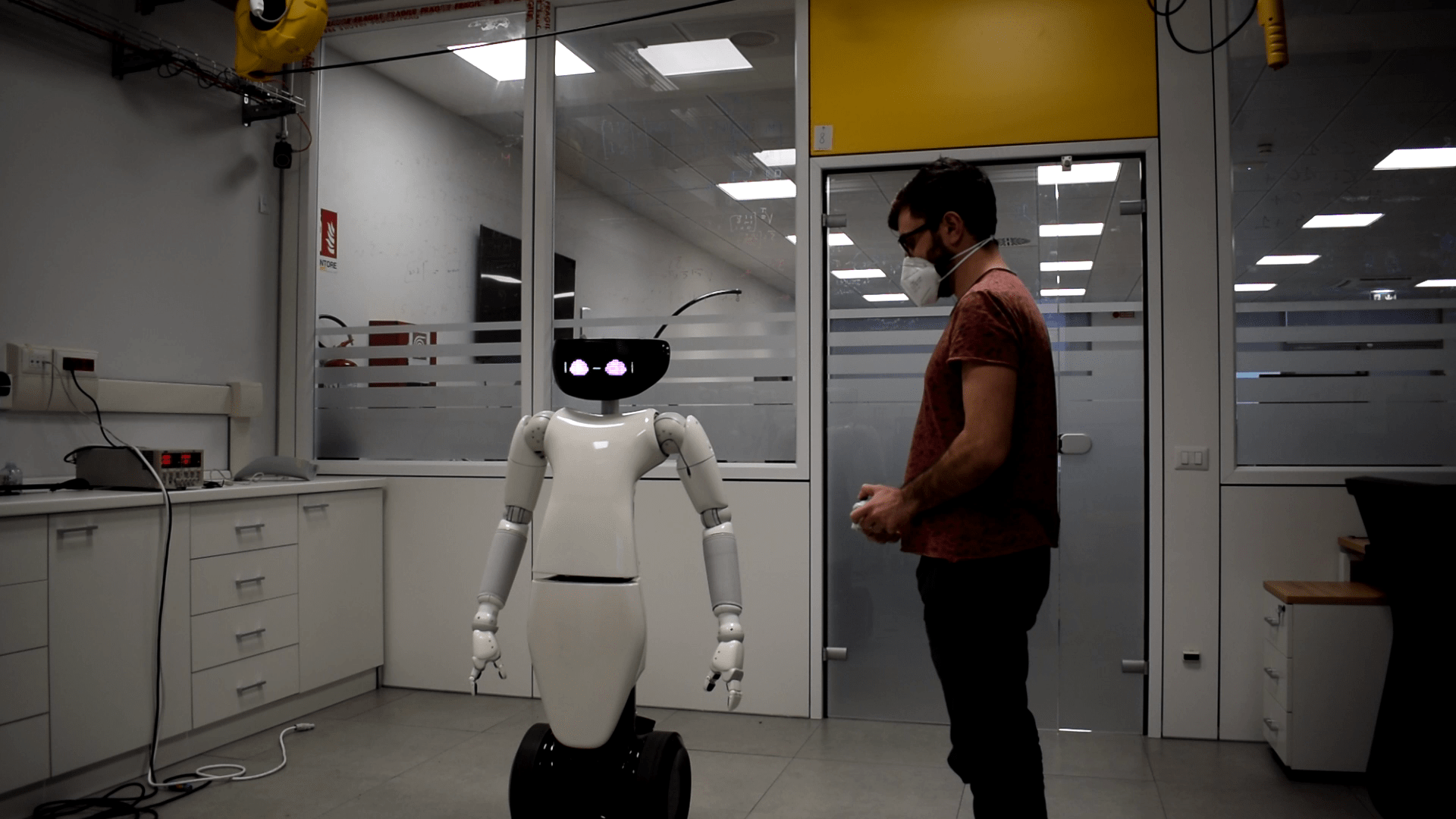}
\caption{A malicious user moves the robot elsewhere. The battery level gets lower than 20\%.}
\label{s2.fig.exec4}
\end{subfigure}
\caption{Execution steps of Experiment~\ref{experiment.robot}.}
\label{s2.fig.exec}

\end{figure}
\vspace{-1.8em}
\begin{figure}[h!]
\centering
\begin{subfigure}[t]{0.45\columnwidth}
\centering
\includegraphics[width=0.8\columnwidth]{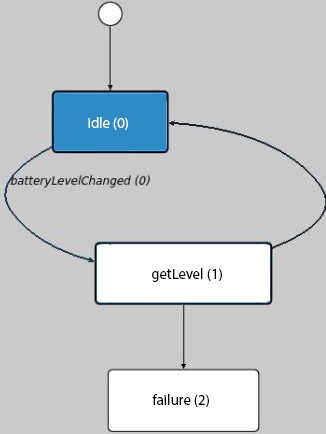}
\caption{Battery level 35\%. The property is satisfied}
\label{s2.fig.mon1}
\end{subfigure}
\hfill
\begin{subfigure}[t]{0.45\columnwidth}
\centering
\includegraphics[width=0.8\columnwidth]{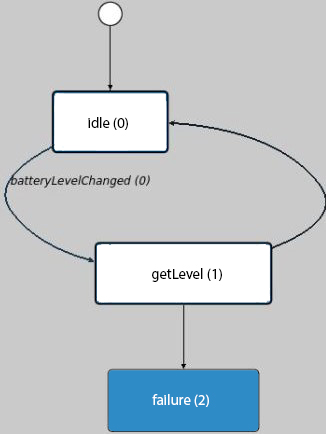}
\caption{Battery level 19\%. The property is violated}
\label{s2.fig.mon2}
\end{subfigure}
\caption{FSM of the runtime monitor for Experiment~\ref{experiment.robot}.}
\label{s2.fig.mon}

\end{figure}
\end{experiment}

%% file: discussion.tex
We formalized the execution context of BTs and
obtained runtime monitors from robot task requirements.  We provided experimental evidence that deliberative level
control can be formalized in a natural yet precise way
and that such formalization can be easily embedded in model-based
design workflows~\cite{ColledanchiseTRO17,biggar2020framework}.
Finally, we implemented our approach on a real robot with BTs at it's  the deliberative level.

The formalization of the execution context and the monitor generation enable the runtime verification of policies in form of BTs. Nevertheless, while the RobMosys framework motivates our choice of using BTs, our approach can be applied to any deliberation model that admits a semantic representation in terms of compositions of state machines.

We can automate our approach by generating the monitors for the formal properties considered in this paper algorithmically. State-of-the-art approaches~\cite{DBLP:conf/memocode/DokhanchiHTF16} suggest that automatic generation can be done in polynomial time.
This allows us to scale our approach efficiently, providing a middleware that can route all the monitored signals efficiently.